\documentclass[twoside,11pt]{article}

\usepackage{jmlr2e}
\usepackage{booktabs}
\usepackage{listings}
\usepackage{amsmath}
\usepackage{comment}
\usepackage{enumitem}
\usepackage{lineno,hyperref}
\usepackage{amssymb}
\usepackage{verbatim}
\usepackage[ruled,vlined]{algorithm2e}
\usepackage{float}
\usepackage{xfrac}
\usepackage[all]{xy}
\usepackage{siunitx}
\usepackage{color}
\usepackage{nth}
\usepackage{todonotes}
\usepackage[ampersand]{easylist}
\usepackage{graphicx}
\usepackage{pgfplots}
\pgfplotsset{compat=1.14}
\pgfplotsset{compat=newest}
\pgfplotsset{plot coordinates/math parser=false}
\usepackage{tikzscale}
\usetikzlibrary{matrix,chains,positioning,decorations.pathreplacing,arrows}
\usepackage{tikz-qtree,tikz-qtree-compat}
\usetikzlibrary{calc}

\ShortHeadings{Kandinsky Patterns}{M\"uller \& Holzinger}
\firstpageno{1}

\begin{document}

\title{Kandinsky Patterns}

\author{\name Heimo M\"uller\textsuperscript{1} \email  heimo.mueller@medunigraz.at\\
\name Andreas Holzinger\textsuperscript{1,2}  \email andreas.holzinger@medunigraz.at\\
\\
\addr \textsuperscript{1} Medical University Graz, Austria\\
\addr \textsuperscript{2} Graz University of Technology, Austria\\
}

\maketitle
\begin{abstract}

Kandinsky Figures and Kandinsky Patterns are mathematically describable, simple self-contained hence controllable test data sets for the development, validation and training of explainability in artificial intelligence. Whilst Kandinsky Patterns have these computationally manageable properties, they are at the same time easily distinguishable from human observers. Consequently, controlled patterns can be described by \textit{both} humans and computers. We define a Kandinsky Pattern as a set of Kandinsky Figures, where for each figure an "infallible authority" defines that the figure belongs to the Kandinsky Pattern.  With this simple principle we build training and validation data sets for automatic interpretability and context learning. In this paper we describe the basic idea and some underlying principles of Kandinsky Patterns and provide a Github repository to invite the international machine learning research community to a challenge to experiment with our Kandinsky Patterns to expand and thus make progress in the field of explainable AI and to contribute to the upcoming field of explainability and causability. 

\end{abstract}
\vspace{0.4cm}

\section{Introduction}
\label{Introduction}

When talking about explainable AI it is important from the very beginning to differentiate between Explainability and Causability: under explainability we understand the property of the AI-system to generate machine explanations, whilst causability is the property of the human to understand the machine explanations \citep{HolzingerEtAl:2019:Wiley-Paper}. Consequently, the key to effective human-AI interaction is an efficient mapping of explainability with causability. Compared to the map metaphor, this is about establishing connections and relations - not drawing a new map. It is about identifying the \textit{same areas in two completely different maps. } When explaining predictions of deep learning models we apply an explanation method, e.g. simple sensitivity analysis, to understand the prediction in terms of the input variables. The result of such an explainability method can be a heatmap. This visualization indicates which pixels need to be changed to make the image look (from the AI-systems perspective!) more or less like the predicted class \citep{SamekWiegandMueller:2017:ExAI}. On the other hand there are the corresponding human concepts and "contextual understanding" needs effective mapping of them both \citep{LakeSalakTenenbaum:2015:ConceptLearning}, and is among the future grand goal of human-centered AI \citep{HolzingerBiemannPattichisKell:2017:Explainable-AI}.  

Wassily Kandinsky (1866--1944) was an influential Russian painter \citep{Duechting:2000:Kandinsky}. As his career progressed, Kandinsky produced increasingly abstract images. For a period from 1922 to 1933 he taught at the famous Bauhaus school in Germany, which celebrated simple colors and forms. Kandinsky was a theorist as well as an artist, and he derived profound meaning from aesthetic experiences. One of Kandinsky's ideas was that there are certain fundamental associations between colors and shapes \citep{Kandinsky:1912:Formfrage}, e.g. he proposed Yellow-Triangle, Blue-Circle, and Red-Square. These associations were formulated introspectively, however, he did conduct his own survey at the Bauhaus in 1923 and postulated a correspondence between color and form. Subsequent empirical studies used preference judgments to test Kandinsky's original color-form combinations, usually yielding inconsistent results. Recent findings suggest that there is no implicit association between the original color-form combinations hence can not considered as a universal property of the visual system \citep{MakinWuerger:2013:KandinskyShape}. In our work we do not pursue this hypothesis any further, but take only the visual principles of Kandinsky as starting point and eponym for the following definitions.

\section{Kandinsky Patterns}
\label{Patterns}

\begin{definition}
A \textbf{Kandinsky Figure} is a square image containing \textit{1} to \textit{n} geometric objects. Each object is characterized by its shape, color, size and position within this square. Objects do not overlap and are not cropped at the border. All objects must be easily recognizable and clearly distinguishable by a human observer. 
\end{definition}

\begin{figure}[h!]
\begin{center}
\includegraphics[width=160pt]{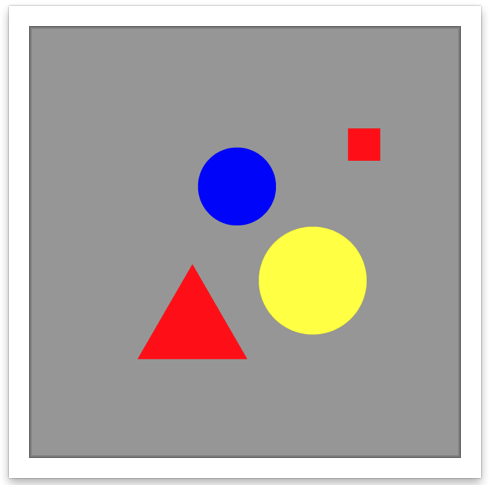}
\end{center}
\caption{A Kandinsky Figure with 4 Objects}
\label{fig:kandinsky-figure}
\end{figure}

The set of all possible Kandinsky Figures $K$ is given by the definition 1 with a specific set of values for shape, color, size, position and the number of geometric objects. In the following examples we use for shape the values circle, square and triangle; for color we use the values red, blue, yellow, and we allow arbitrary positions and size with the restriction that it is still recognizable. Furthermore, we require each Kandinsky Figure to contain exactly 4 objects in the following illustrative examples. In the demo implementation this fact is embedded in the base class "Kandinsky Universe", and in the generator functions\footnote{\url{https://github.com/human-centered-ai-lab/app-kandinsky-pattern-generator}}, see Figure \ref{fig:kandinsky-figure}. 

\begin{definition} A Statement \textbf{$s(k)$} about a Kandinsky Figure $k$ is either a mathematical function, $s(k) \to B$; with $B (0,1)$ or a natural language statement, which is either true or false. 
\end{definition}

Remark: The evaluation of a natural language statement is always done in a specific \textit{context}. In the followings examples we use well known concepts from human perception and linguistic theory. If {$s(k)$} is given as an algorithm, it is essential that the function is a pure function, which is a computational analogue of a mathematical function. 

\begin{definition}
A \textbf{Kandinsky Pattern} $K$ is defined as the subset of all possible Kandinsky Figures $k$ with $s(k) \to 1$ or the natural language statement is true.  $s(k)$ and a natural language statement are equivalent, if and only if the resulting Kandinsky Patterns contains the same Kandinsky Figures. $s(k)$ and the natural language statement are defined as the \textbf{Ground Truth} of a Kandinsky Pattern. 
\end{definition}

\begin{figure}[h!]
\begin{center}
\includegraphics[width=0.8\textwidth]{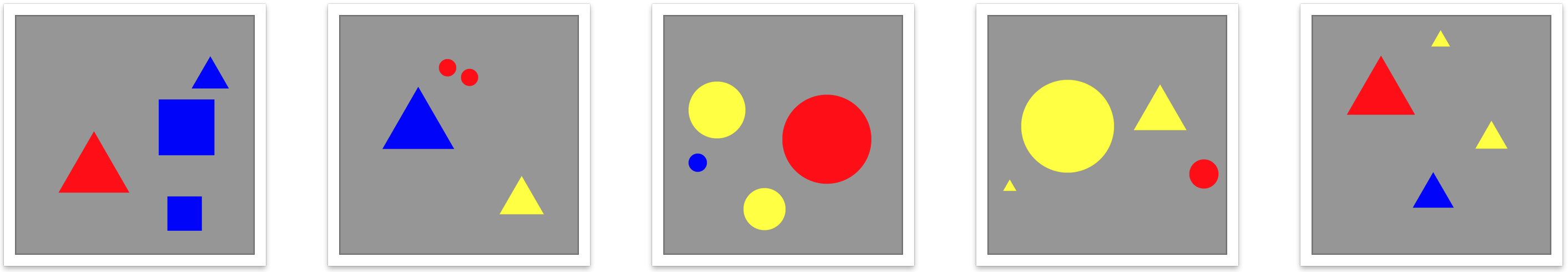}
\end{center}
\caption{Five Kandinsky Figures of a Kandinsky Pattern}
\label{fig:kandinsky-pattern}
\end{figure}

Figure \ref{fig:kandinsky-pattern} shows an example of a Kandinsky Pattern with the natural language statement: \textit{"the Kandinsky Figure has two pairs of objects with the same shape, in one pair the objects have the same color, in the other pair different colors, two pairs are always disjunct, i.e. they don't share a object"}. In machine learning such in classification algorithms it is usually not given such a simple function, but it is given as a highly non-linear, high-dimensional network.  The aim of explanation in machine learning is to identify areas of activation within the network structure which correspond to concepts in the natural language statement.

\textbf{Problem 1}: How can we explain a Kandinsky Pattern, if we do not know the Ground Truth and the membership of Kandinsky Figures to a Kandinsky Pattern is only known for a limited number of Kandinsky Figures. 

\textbf{Problem 2:} Generate a natural language statement, which is easy understandable and equivalent to the machine explanation (classification algorithm).

The process of explanation is the generation and refinement of a hypothesis to find the underlying description. The validation is achieved by the scientific method of asking a question, forming a testable hypothesis, setting up the experimental design, running the experiment and either accepting the hypothesis, rejecting it or, in the third case according to \citep{Popper:1935:Logik}, one cannot make any assumption.

The ground truth is used to prove or disprove research hypotheses. "Ground truthing" consequently refers to the process of collecting the proper objective (provable!) data for testing the hypothesis. For a machine learning algorithm an explanation can be seen as the successful classification algorithm of a Kandinsky pattern. 

The following example illustrates the above: 
\newpage
The ground truth $gt(k) =$  "\textit{the Kandinsky Figure has two pairs of objects with the same shape, in one pair the objects have the same colors in the other pair different colors, two pairs are always disjunct, i.e. they don't share  objects"} defines the Kandinsky Pattern $K_{gt}$, see Figure \ref{fig:kandinsky-pattern}. 

\begin{figure}[h!]
\begin{center}
\includegraphics[width=0.8\textwidth]{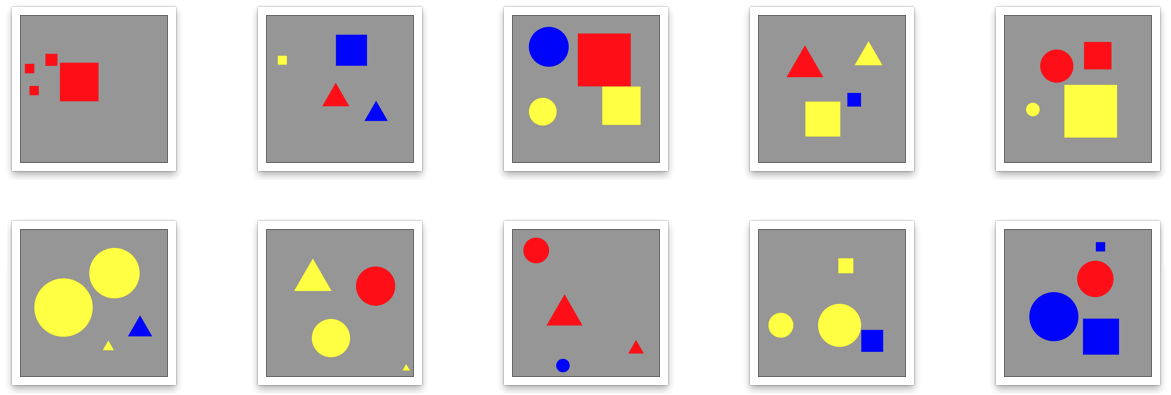}
\end{center}
\caption{Kandinsky Figures of $h_1(k)$, the first row shows  counterfactuals }
\label{fig:KF-h1}
\end{figure}

For a more general hypothesis $h_1(k)$ = \textit{"the Kandinsky Figure has two pairs of objects with the same shape"} we see that $K_{h1} \setminus K_{gt} \neq \emptyset$, i.e. the Kandinsky Pattern of  $h_1(k)$ contains Kandinsky Figures which are not in the Kandinsky Pattern of the ground truth. Figure \ref{fig:KF-h1} shows Kandinsky Figures according to  $h_1(k)$, the first row is a contradiction to the ground truth, i.e. it falsifies $h_1(k)$.

\begin{figure}[h!]
\begin{center}
\includegraphics[width=0.8\textwidth]{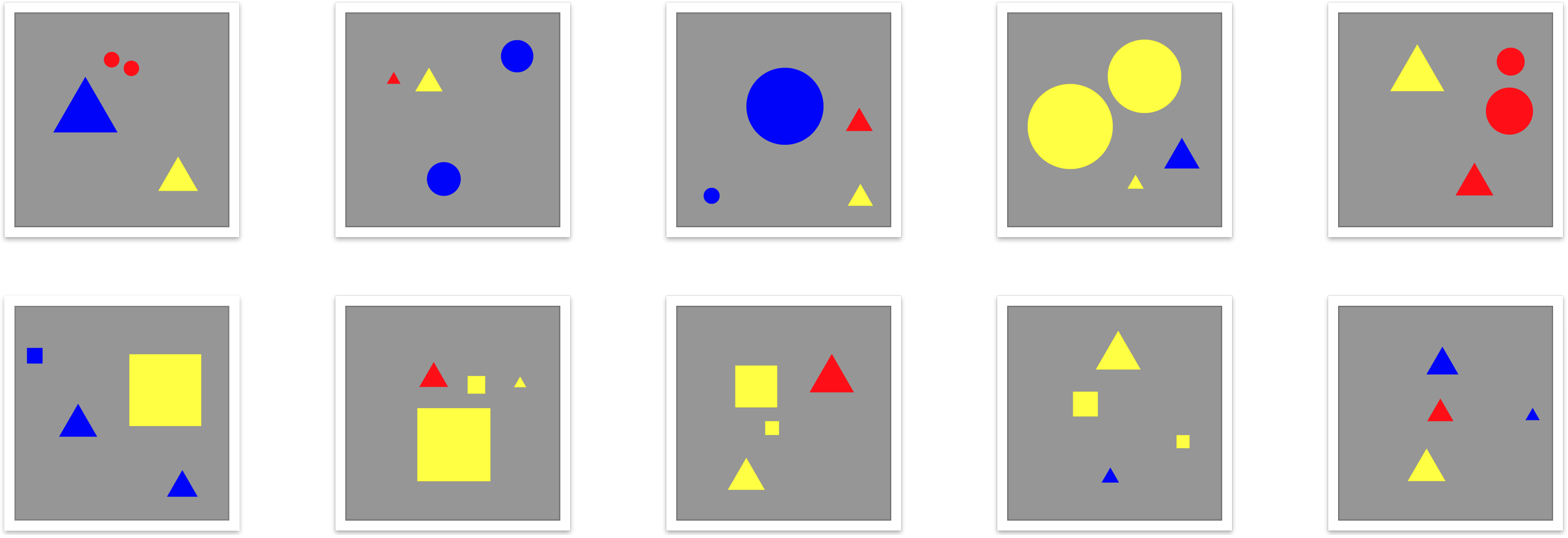}
\end{center}
\caption{Kandinsky Figures of $h_2(k)$ in the first row, Kandinsky Figures from $K_{gt}$ in the second row, which falsify $h_2(k)$.}
\label{fig:KF-h2}
\end{figure}

A specific hypothesis as $h_2(k)$ = \textit{"the Kandinsky Figure consist of two triangles with different color and two circles of  same color"} generates a Kandinsky Pattern $K_{h2}$ with $K_{gt}  \setminus K_{h2}  \neq \emptyset$, i.e. the Kandinsky Pattern of  $h_2(k)$ is missing Kandinsky Figures which are in the Kandinsky Pattern of the ground truth. Figure \ref{fig:KF-h2} shows in the first row Kandinsky Figures according to  $h_2(k)$ and Kandinsky Figures from $K_{gt}$ in the second row, which falsify  $h_2(k)$. $K_{h} \setminus K_{gt} \cup  K_{gt}\setminus K_{h}$ is the set of counterfactual Kandinsky Figures for a given hypothesis $h$.

\section{Background}

In a natural language statement about a Kandinsky Figure humans use a series of basic concepts which are combined through logical operators. The following (incomplete) examples illustrate some concepts of increasing complexity.

\begin{itemize}[noitemsep]
  \item Basic concepts given by the definition of a Kandinsky Figure: a set of \textit{objects}, described by \textit{shape}, \textit{color},  \textit{size} and \textit{position}.
  \item Existence, numbers, set-relations (\textit{number}, \textit{quantity} or \textit{quantity ratios} of objects), e.g. \textit{"a Kandinsky Figure contains 4 red triangles and more yellow objects than circles". }
  \item Spatial concepts describing the arrangement of objects, either absolute (\textit{upper}, \textit{lower}, \textit{left}, \textit{right}, \dots) or relative (\textit{below}, \textit{above}, \textit{on top}, \textit{touching}, \dots), e.g. \textit{"in a Kandinsky Figure red objects are on the left side, blue objects on the right side, and yellow objects are below blue squares".}
  \item Gestalt concepts (see below) e.g. \textit{closure}, \textit{symmetry}, \textit{continuity}, \textit{proximity}, \textit{similarity}, e.g. \textit{"in a Kandinsky Figure objects are grouped in a circular manner".} 
  \item Domain concepts, e.g. \textit{"a group of objects is perceived as a "flower"".}
\end{itemize}

In their experiments \citep{HubelWiesel:1962:CatVisual} discovered, among others, that the visual system builds an image from very simple stimuli into more complex representations. This inspired the neural network community to see their so-called "deep learning" models as a cascading model of cell types, which follows always similar simple rules: at first lines are learned, then shapes, then objects are formed, eventually leading to \textbf{concept representations.} By use of backpropagation such a model is able to discover intricate structures in large data sets to indicate how the internal parameters should be adapted, which are used to compute the representation in each layer from the representation in the previous layer \citep{LeCunBengioHinton:2015:DeepLearningNature}. Building \textit{concept representations} refers to the human ability to learn categories for objects and to recognize new instances of those categories. In machine learning, concept learning is defined as the inference of a Boolean-valued function from training examples of its inputs and outputs \citep{Mitchell:1997:MachineLearningBook} in other words it is training an algorithm to distinguish between examples and non-examples (we call the latter counterfactuals). 

\textbf{Concept learning} has been a relevant research area in machine learning for a long time and had it origins in cognitive science, defined as search for attributes which can be used to distinguish exemplars from non exemplars of various categories \citep{Bruner:1956:InStudyOfThinking}. The ability to think in abstractions is one of the most powerful tools humans possess. Technically, humans order their experience into coherent categories by defining a given situation as a member of that collection of situations for which responses \textit{x, y,} etc. are most likely appropriate. This classification is not a passive process and to understand how humans learn abstractions is essential not only to the understanding of human thought, but to building artificial intelligence machines \citep{Hunt:1962:conceptLearning}. 

In computer vision an important task is to find a likely interpretation \textit{W} for an observed image \textit{I}, where \textit{W} includes information about the spatial location, the extent of objects, the boundaries etc. Let \textit{SW} be a function associated with a interpretation \textit{W} that encodes the spatial location and extent of a component of interest, where $SW_{(i,j)}$ = 1 for each image location $(i,j)$ that belongs to the component and 0 else-where. Given an image, obtaining an optimal or even likely interpretation \textit{W}, or associated \textit{SW}, can be difficult. For example, in edge detection previous work \citep{DollarEtAl:2006:SupervisedEdgeLearning} asked what is the probability of a given location in a given image belonging to the component of interest. 

\cite{Tenenbaum:1999:HumanConcept} presented a model of concept learning that is both computationally grounded and able to fit to human behaviour. He argued that two apparently distinct modes of generalizing concepts -- abstracting rules and computing similarity to exemplars -- should both be seen as special cases of a more general \textit{Bayesian learning framework}. Originally, Bayes (and more specific \citep{Laplace:1781:MemoireProbabilites}) explained the specific workings of these two modes, i.e. which rules are abstracted, how similarity is measured, why generalization should appear in different situations. This analysis also suggests why the rules/similarity distinction, even if not computationally fundamental, may still be useful at the algorithmic level as part of a principled approximation to fully Bayesian learning.

\textbf{Gestalt-Principles} ("Gestalt" = German for shape) are a set of empirical laws describing how humans gain meaningful perceptions and make sense of chaotic stimuli of the real-world. As Gestalt-cues they have been used in machine learning for a long time. Particularly, in learning classification models for segmentation, the task is to classify between "good" segmentations and "bad" segmentations and to use the Gestalt-cues as features (the priors) to train the learning model. Images segmented manually by humans are used as examples of "good" segmentations (ground truth), and "bad" segmentations are constructed by randomly matching a human segmentation to a different image \citep{RenMalik:2003:GestaltMachine}. Gestalt-principles \citep{Koffka:1935:Gestalt} can be seen as rules, i.e. they discriminate competing segmentations only when everything else is equal, therefore we speak more generally as Gestalt-laws and one particular group of Gestalt-laws are the Gestalt-laws of grouping, called Pr\"agnanz \citep{Wertheimer:1938:GestaltLaws}, which include the law of Proximity: objects that are close to one another appear to form groups, even if they are completely different, the Law of Similarity: similar objects are grouped together; or the law of Closure: objects can be perceived as such, even if they are incomplete or hidden by other objects.

Unfortunately, the currently best performing machine learning methods have a number of disadvantages, and one is of particular relevance: Neural networks ("deep learning") are difficult to interpret due to their complexity and are therefore considered as "black-box" models \citep{HolzingerEtAl:2017:glassbox}. Image Classifiers operate on low-level features (e.g. lines, circles, etc.) rather than high-level concepts, and with domain concepts (e.g images with a storefront). This makes their inner workings difficult to interpret and understand. However, the "why" would often be much more useful than the simple classification result. 

\section{Related Work}

Reasoning and explanation has a long history within the AI/machine learning community \cite{PooleMackworthGoebel:1998:CompIntelligence} and recently quite a number of authors proposed mechanisms for generating explanations by deep learning models. In the following we can present only a tiny fraction of related work and apologize for any work not mentioned here.

Kandinsky Patterns can be used as validation data set for experiments in explainability, similarly as in the following works: \cite{HendricksAkataEtAl:2016:VisualExplanations}  proposed a model that focused on discriminating properties of a visible object and jointly predicts a class label. They explained why the predicted label is appropriate for the respective image on the basis of a loss function based on sampling and reinforcement learning that learns to generate sentences that realize a global sentence property, such as class specificity. 
\cite{SelvarajuEtAl:2017:GradCAM}  proposed a technique for producing ‘visual explanations’ following Gradient-weighted Class Activation Mapping (Grad-CAM), which uses the gradients of any target concept (e.g. logits for "dog" or a caption, etc.), influencing the final convolutional layer to produce a coarse localization map to highlight relevant regions in the image for predicting the concept.
\cite{KimEtAl:2017:TCAV}  introduced so called Concept Activation Vectors (CAVs), which provide an interpretation of a neural networks internal state in terms of more human-friendly concepts. Their key idea is to view the high-dimensional internal state of a neural network as an aid and not as an obstacle. Technically, they use directional derivatives to quantify the degree to which a user-defined concept is important to a classification result. That means e.g., in a classifier, how sensitive a prediction of a certain area is for a certain concept. 
\cite{JohnsonEtAl:2017:Clevr} presented a different approach,
called CLEVR, which contains a diagnostic data set  for testing visual reasoning abilities. The code can be used to render synthetic images and compositional questions for those images e.g. \textit{"How many small spheres are there ?".} Each question in CLEVR is represented both in natural language and as a functional program, the latter representation allows for precise determination of the reasoning skills required to answer each question. Questions in CLEVR test various aspects of visual reasoning including attribute identification, counting, comparison, spatial relationships, and logical operations. However, this work does not deal with concept learning. 
\cite{MaoEtAl:2015:LearningChild} described a similar approach already earlier by addressing the task of learning novel visual concepts and their interactions with other concepts from a few images with sentence descriptions: for each word in a sentence, their model takes the current word index and the image as inputs and outputs the next word index.

\section{Data Sets and Challenges}
\label{datasets}

Kandinsky Patterns can be used as test data sets for various research questions, e.g. to address and evaluate the following topics:
\begin{enumerate}[noitemsep]
   \item Describe classes of Kandinsky Patterns according to their ability to be classified by machine learning algorithms in comparison to human explanation strategies. 
   \item Investigate transfer learning of concepts as numbers, geometric positions and Gestalt principles in the classification and explanation of Kandinsky Patterns.
   \item Develop mapping strategies from an algorithmic classification to a known human explanation of a Kandinsky Pattern.
   \item Automatic generation of a human understandable explanation of a Kandinsky Pattern.
    
\end{enumerate}

We invite the international machine learning community to experiment with our Kandinsky data set\footnote{\url{https://github.com/human-centered-ai-lab/dat-kandinsky-patterns}}, and re-use and contribute to the Kandinsky software tools\footnote{\url{https://github.com/human-centered-ai-lab/app-kandinsky-pattern-generator}}.

Please note that the main aim of the training data sets and the following challenges is not in the evaluation of machine learning algorithms, but most of all \textit{in explaining the successful classification by human understandable statements. }


\subsection{Challenge 1 - Objects and Shapes}

In the challenge \textbf{Objects and Shapes} the ground truth gt(k) is defined as "\textit{in a Kandinsky Figure small objects are arranged on big shapes same as  object shapes, in the big shape of type X, no small object of type X exists. Big square shapes only contain blue and red objects, big triangle shapes only contain yellow and red objects and big circle shape contain only yellow and blue objects"}. 

\begin{figure}[h!]
\begin{center}
\includegraphics[width=0.8\textwidth]{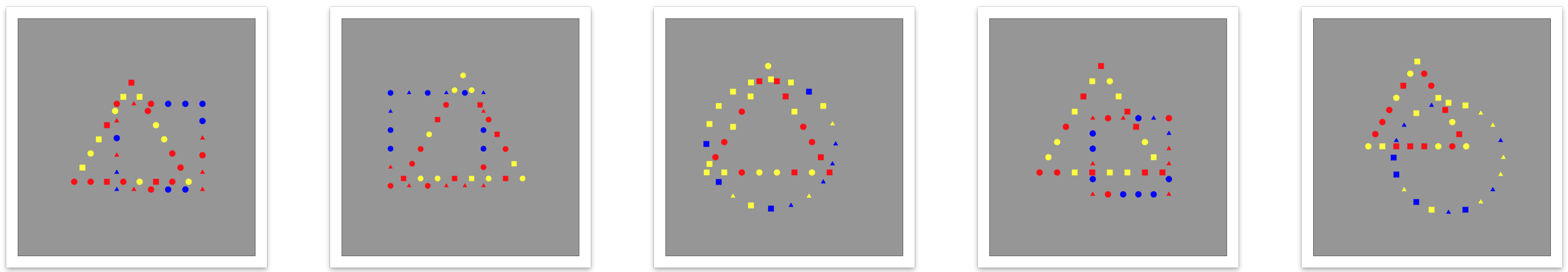}
\end{center}
\caption{Kandinsky Figures according to ground truth of challenge 1 }
\label{fig:kandinsky-challenge-1-true}
\end{figure}

\begin{figure}[h!]
\begin{center}
\includegraphics[width=0.8\textwidth]{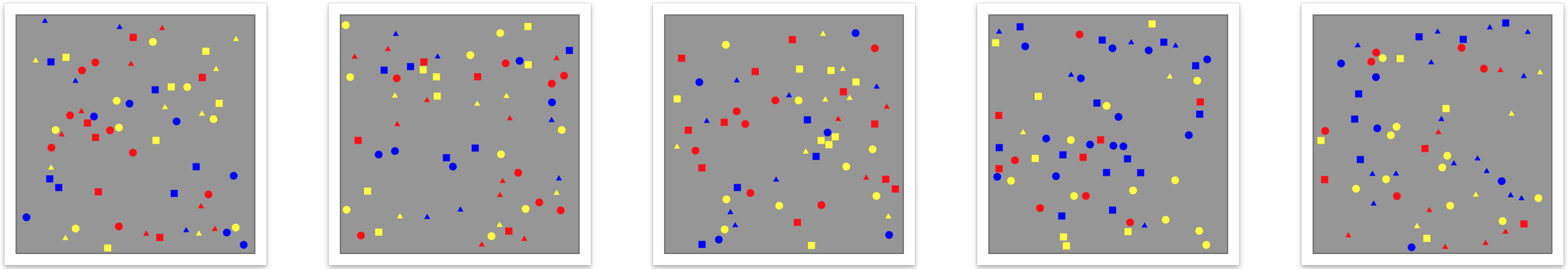}
\end{center}
\caption{Kandinsky Figures not belonging to the Kandinsky Pattern of challenge 1}
\label{fig:kandinsky-challenge-1-false}
\end{figure}

\begin{figure}[h!]
\begin{center}
\includegraphics[width=0.8\textwidth]{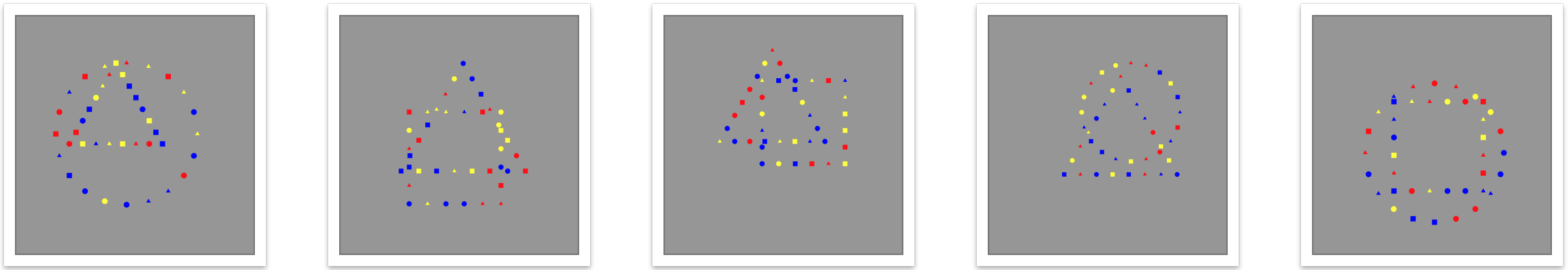}
\end{center}
\caption{Kandinsky Figures which falsify a simple hypothesis for challenge 1 }
\label{fig:kandinsky-challenge-1-counterfactual}
\end{figure}

Figure \ref{fig:kandinsky-challenge-1-true} shows Kandinsky Figure according to ground truth. Figure \ref{fig:kandinsky-challenge-1-false} shows random Kandinsky Figure with approximately the same number of objects not belonging to the Kandinsky Pattern and Figure \ref{fig:kandinsky-challenge-1-counterfactual} Kandinsky Figures which are generated with a simple but not valid hypothesis. 

\begin{itemize}[noitemsep]
\item \textbf{Question 1}: \textit{Which machine learning algorithm can classify Kandinsky Figures of challenge 1.} 
\item \textbf{Question 2:} \textit{Identify layers and regions in the network, which corresponds to "small" and "big" shapes and the restrictions on object membership and color.}
\end{itemize}

Download the data set for challenge 1 here: \url{https://tinyurl.com/Kandinsky-C1}\footnote{\url{https://github.com/human-centered-ai-lab/dat-kandinsky-patterns/tree/master/challenge-nr-1}}

\subsection{Challenge 2 - Nine Circles}

\begin{figure}[h!]
\begin{center}
\includegraphics[width=0.8\textwidth]{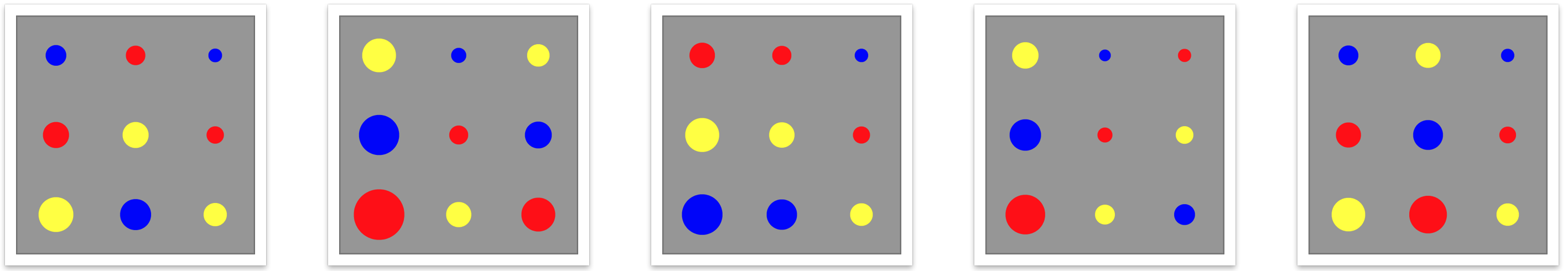}
\end{center}
\caption{Kandinsky Figures according to ground truth of challenge 2 }
\label{fig:kandinsky-challenge-2-true}
\end{figure}

\begin{figure}[h!]
\begin{center}
\includegraphics[width=0.8\textwidth]{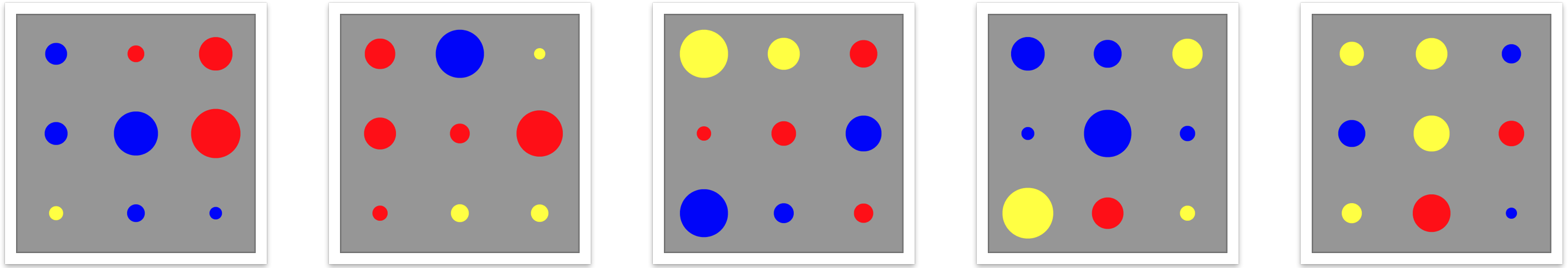}
\end{center}
\caption{Kandinsky Figures not belonging to the Kandinsky Pattern of challenge 2}
\label{fig:kandinsky-challenge-2-false}
\end{figure}

\begin{figure}[h!]
\begin{center}
\includegraphics[width=0.8\textwidth]{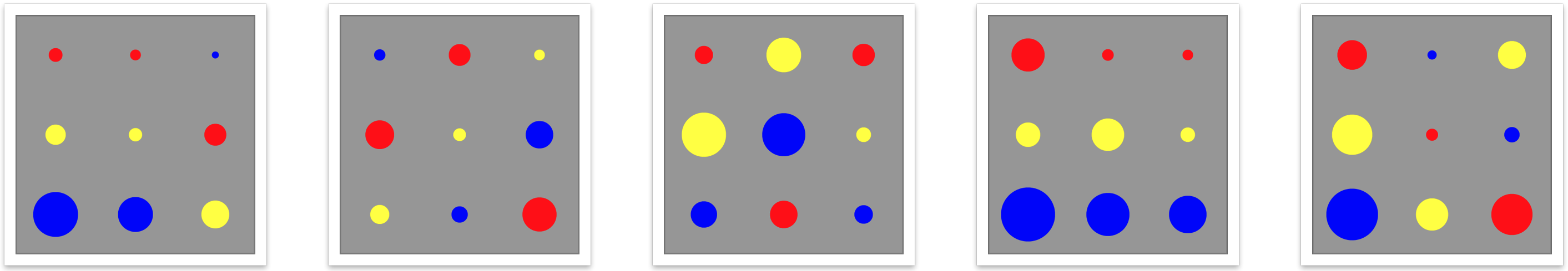}
\end{center}
\caption{Kandinsky Figures which falsify a simple hypothesis for challenge 2 }
\label{fig:kandinsky-challenge-2-counterfactual}
\end{figure}

In the challenge \textbf{Nine Circles} the set of Kandinsky Figures consist of 9 circles arranged in a regular grid. Figure \ref{fig:kandinsky-challenge-2-true} shows Kandinsky Figure according to ground truth. Figure \ref{fig:kandinsky-challenge-2-false} shows Kandinsky Figures not belonging to the Kandinsky Pattern and Figure \ref{fig:kandinsky-challenge-2-counterfactual} shows Kandinsky Figures which are "almost true", i.e. they fulfill a hypothesis similar to ground truth, but are counter factual. 

\begin{itemize}[noitemsep]
\item \textbf{Question 1}: \textit{Explain the Kandinsky Pattern in an algorithmic way, i.e. train a network which classifies Kandinsky Figures according to ground truth of challenge 2.}
\item \textbf{Question 2:} \textit{Explain the Kandinsky Pattern in natural language.}
\end{itemize}

Download the data set for challenge 2 here: \url{https://tinyurl.com/Kandinsky-C2}\footnote{\url{https://github.com/human-centered-ai-lab/dat-kandinsky-patterns/tree/master/challenge-nr-2}} 

\vspace{1cm}
\subsection{Challenge 3 - Blue and Yellow Circles}

\begin{figure}[h!]
\begin{center}
\includegraphics[width=0.8\textwidth]{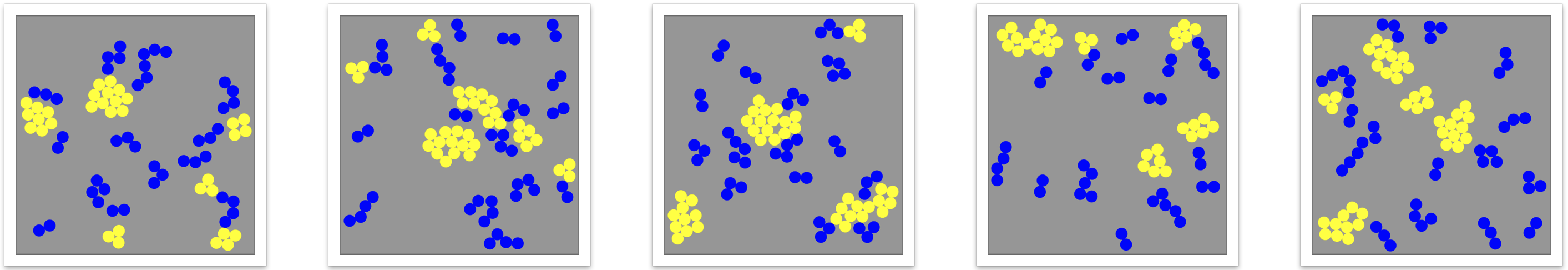}
\end{center}
\caption{Kandinsky Figures according to ground truth of challenge 3 }
\label{fig:kandinsky-challenge-3-true}
\end{figure}

\begin{figure}[h!]
\begin{center}
\includegraphics[width=0.8\textwidth]{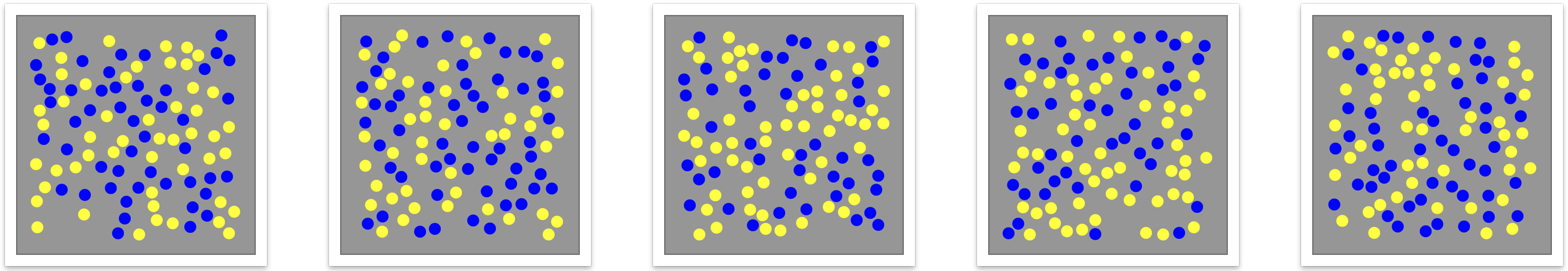}
\end{center}
\caption{Kandinsky Figures not belonging to the Kandinsky Pattern of challenge 3}
\label{fig:kandinsky-challenge-3-false}
\end{figure}

\begin{figure}[h!]
\begin{center}
\includegraphics[width=0.8\textwidth]{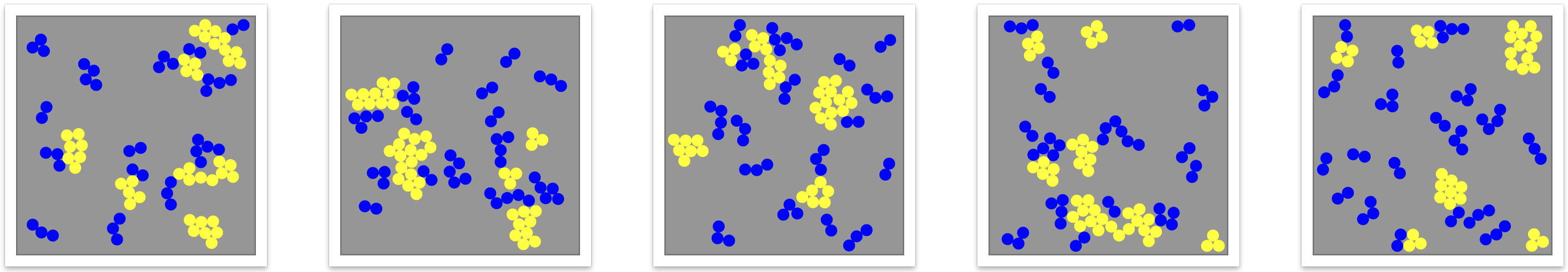}
\end{center}
\caption{Kandinsky Figures which falsify a simple hypothesis for challenge 3}
\label{fig:kandinsky-challenge-3-counterfactual}
\end{figure}

In the challenge \textbf{ Blue and Yellow Circles} the set  of all possible Kandinsky Figures consist of equal size  blue and yellow circles. Figure \ref{fig:kandinsky-challenge-3-true} shows Kandinsky Figures according to ground truth. Figure \ref{fig:kandinsky-challenge-3-false} shows Kandinsky Figures with approximately the same number of objects not belonging to the Kandinsky Pattern and Figure \ref{fig:kandinsky-challenge-3-counterfactual} Kandinsky Figures which are "almost true", i.e. they fulfill a hypothesis similar to the ground truth, but are counter factual. 

\begin{itemize}[noitemsep]
\item \textbf{Question 1}: \textit{Explain the Kandinsky Pattern in an algorithmic way, i.e. train a network which classifies Kandinsky Figures according to ground truth of challenge 3.}
\item \textbf{Question 2:} \textit{Explain the Kandinsky Pattern in natural language.}
\end{itemize}

Download the data set for challenge 3 here: \url{https://tinyurl.com/Kandinsky-C3} \footnote{\url{https://github.com/human-centered-ai-lab/dat-kandinsky-patterns/tree/master/challenge-nr-3}} 

\newpage
\section{Conclusion}

By comparing both the strengths of machine intelligence and human intelligence it is possible to solve problems where we are currently lacking appropriate methods. One grand general question is "How can we perform a task by exploiting knowledge extracted during solving previous tasks?" To answer this question it is necessary to get insight into human behavior, but not with the goal of mimicking human behavior, rather to contrast human learning methods to machine learning methods. We hope that our Kandinsky Patterns challenge the international machine learning community and we are looking forward to receiving comments and results. Updated information can be found at the accompanying Web page\footnote{\url{https://human-centered.ai/kandinksy-challenge}}.

\section{Acknowledgements}

We are grateful for interesting discussions with our local and international colleagues and their encouragement.

\bibliography{references}

\end{document}